\documentclass[journal,twoside,print]{ieeecolor}
\usepackage{generic}
\usepackage{cite}
\usepackage{amsmath,amssymb,amsfonts}
\usepackage{algorithmic}
\usepackage{graphicx}
\usepackage{textcomp}

\usepackage{graphicx}

\usepackage{subfig}
\usepackage{overpic}

\usepackage{multirow}
\usepackage{booktabs}
\usepackage{threeparttable}
\usepackage{cite}
\usepackage{color}

\def\BibTeX{{\rm B\kern-.05em{\sc i\kern-.025em b}\kern-.08em
    T\kern-.1667em\lower.7ex\hbox{E}\kern-.125emX}}
\begin{document}
\title{Visual Fault Detection of Multi-scale Key Components in Freight Trains}

\author{Yang~Zhang,~Yang~Zhou,~Huilin~Pan,~Bo~Wu,~and~Guodong~Sun
\thanks{Manuscript received ** **, 2022; revised ** **, 2022. accepted ** **, 2022.
Date of publication ** **, 2022; date of current version ** **, 2022. 
\emph{(Corresponding author: Guodong~Sun.)}}
\thanks{Y.~Zhang,~Y.~Zhou,~H.~Pan,~and~G.~Sun are with the School of Mechanical Engineering, Hubei University of Technology, and the Hubei Key Laboratory of Modern Manufacturing Quality Engineering, Wuhan 430068, China (e-mail: yzhangcst@hbut.edu.cn; z-g99@hbut.edu.cn; hlp@hbut.edu.cn; sgdeagle@163.com).}
\thanks{B.~Wu is with the Shanghai Advanced Research Institute, Chinese Academy
of Sciences, Shanghai 201210, China (e-mail: wubo@sari.ac.cn).}
\thanks{Y. Zhang is also with the National Key Laboratory for Novel Software
Technology, Nanjing University, Nanjing 210023, China (email:
yzhangcst@smail.nju.edu.cn).}
\thanks{Color versions of one or more of the figures in this article are 
available online at http://ieeexplore.ieee.org.}
\thanks{Digital Object Identifier}
}

\maketitle

\begin{abstract}
    Fault detection for key components in the braking system of freight trains is critical for ensuring railway transportation safety. Despite the frequently employed methods based on deep learning, these fault detectors are extremely reliant on hardware resources and complex to implement. In addition, no train fault detectors consider the drop in accuracy induced by scale variation of fault parts. This paper proposes a lightweight anchor-free framework to solve the above problems. Specifically, to reduce the amount of computation and model size, we introduce a lightweight backbone and adopt an anchor-free method for localization and regression. To improve detection accuracy for multi-scale parts, we design a feature pyramid network to generate rectangular layers of different sizes to map parts with similar aspect ratios. Experiments on four fault datasets show that our framework achieves 98.44\% accuracy while the model size is only 22.5 MB, outperforming state-of-the-art detectors.
  \end{abstract}
  
  \begin{IEEEkeywords}
  fault detection, freight train images, multi-scale, anchor-free, convolutional neural network.
  \end{IEEEkeywords}

  \section{Introduction}

  \IEEEPARstart{A}{s} a crucial mode of transportation in the country, railways have attracted much attention for their safety and dependability.
  In particular, the reliability of the braking system is vital to the safe operation of trains, as the failure of some key components in the braking system can result in serious consequences.
  Nevertheless, the faults of those key components are usually detected manually. Missing and false detection results are prone to occur due to subjective influence. With the increasing development of visual detection technology in recent years, many academics recommend using machines to replace the manual to detect the faults of critical components.
  For example, Chen et al.~\cite{TII-1} proposed a convolutional neural network (CNN) based method to detect the failure of bogies, which is independent of expertise or prior knowledge.
  Xin et al.~\cite{TII-2} proposed a self-calibrated residual network for detecting wheelset bearing faults.
  Zhang et al.~\cite{TII} designed a new backbone to ensure detection accuracy while significantly reducing the model size.
  
\begin{figure*}[t]
    \begin{center}
        \centering
        \includegraphics[width=7in]{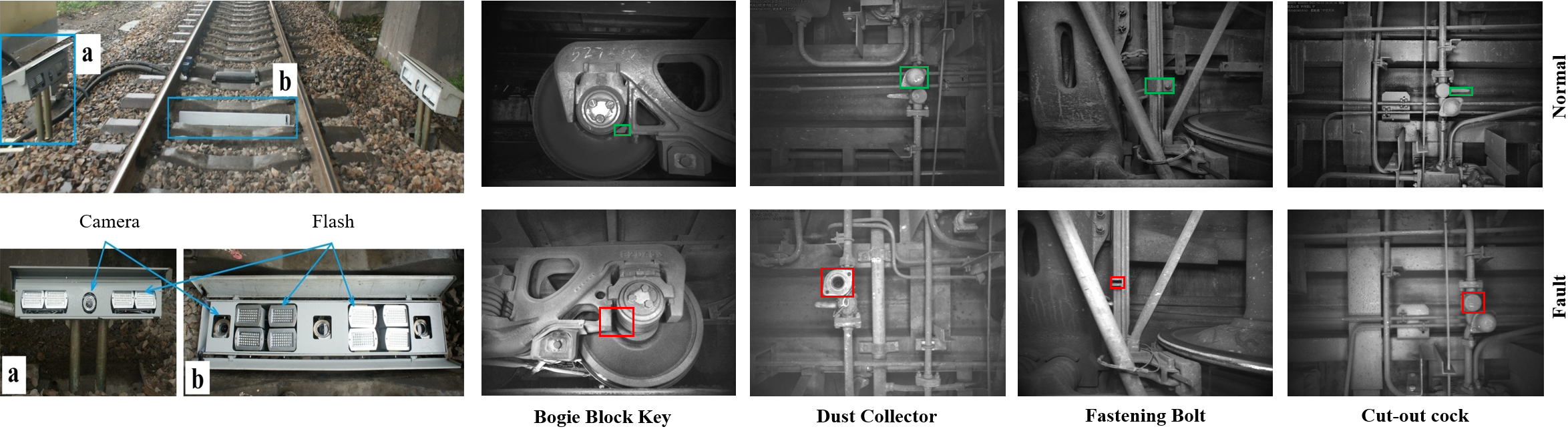}
        \caption{Hardware system for train fault detection in the actual scene, and comparison of normal and fault images in the four datasets. It can be seen that different parts have large differences in scale and aspect ratio.}
        \label{exp}
    \end{center}
\end{figure*}

  Because the fault detection equipment for trains is arranged in the field environment, the computing and hardware resources are limited. Additionally, as illustrated in Fig.~\ref{exp}, the bottom of the train is made up of many different pieces, individually with different dimensions and length-to-width ratios. During detection, ordinary networks struggle to map pieces with varying aspect ratios and sizes.
  As a result of the oversampling of small scale parts and the undersampling of large scale parts, accuracy is substantially affected. 
  Therefore, detecting multi-scale key components in trains while reducing the network dependence on hardware resources is vital to improve detection accuracy and deploy the network in realistic scenarios.
  Specifically, although the two-stage detectors are widely used in object detection tasks, an additional region proposal network (RPN) is required to provide interest regions, resulting in larger model size and computational complexity. So we introduce a one-stage detector for fault detection.
  Inspired by MatrixNets~\cite{xNets}, we consider using the matrix networks to solve the detection problem of multi-scale parts. By generating rectangle layers of different sizes to map objects of similar scales, we can address the problem of scale changes. However, the network creates extra computations while producing those layers, increasing network parameters and model size.

  To address the above problems, we propose a lightweight anchor-free framework for fault detection of multi-scale key components in trains. 
  Specifically, to eliminate calculations related to anchor boxes, we discard the usual concept of bounding boxes and instead employ corners for localization and regression. Moreover, a lightweight backbone is used to reduce the model size further and ensure effective feature extraction capabilities. Finally, we add bottleneck structures and attention modules into the matrix networks, named multi-scale feature pyramid.
  Nevertheless, we sensibly construct the layer ranges and the number of channels of the multi-scale feature pyramid based on the fault characteristics of the critical components in trains. Extensive experiments on four fault datasets show that our framework can be used to detect faults in multi-scale critical components of trains effectively. Compared with state-of-the-art detectors, our framework delivers high accuracy and low resource requirements with a model size of 22.5 MB.  
  
  In general, our main contributions are as follows.
  \begin{enumerate}
      \item{We propose a lightweight anchor-free framework for fault detection of multi-scale critical components in freight trains.
      }
      \item{We use a lightweight but effective backbone and multiscale feature pyramid network for reducing model size while improving accuracy.}
      \item{The experimental results verify that our method is robust to scale changes and achieves the highest detection accuracy under the smallest model size compared with the state-of-the-art methods.}
  \end{enumerate}
  
  The rest of the article is structured as follows. Section~\ref{Related_Works} introduces related works about fault detection for freight train images and anchor-free detectors. Section~\ref{Methods} introduces our framework and various components. To verify the effectiveness of our method, we conduct ablation studies in Section~\ref{EXPERIMENTS}. Finally, Section~\ref{Results} summarizes the full article.
  
  \section{RELATED WORKS}
  \label{Related_Works}
  \subsection{Fault Detection of Freight Train Images}
  Sun et al.~\cite{Sun} proposed an automatic fault detection system to detect four types of faults. The detection accuracy of the system was higher than 92.5\% in four different fault regions, but the detection results were susceptible to light interference with low robustness.
  Ling et al.~\cite{Ling} proposed a model based on hierarchical features for instance-level prediction of rough defect regions. 
  Chang et al.~\cite{Chang} proposed a method for fault diagnosis and localization of critical small parts of trains and had robustness in undesirable environments such as low texture and high light.
  A fast adaptive Markov random field (FAMRF) algorithm and an exact height function (EHF)~\cite{OE} were used to detect faults in the braking system, which also compared the performance with a cascade detector based on local binary patterns (LBPs).
  Ye et al.~\cite{Ye} proposed a network of multi-feature fusion to detect three typical faults. The detection accuracy reached 88.72\% and had good robustness to complex noise environments.
  Zhang et al.~\cite{Zhang2021} proposed a fault detection method to detect rod springs of fixtures with an experimental accuracy of 91.98\%.
  Zhou et al.~\cite{Zhou} proposed a detection algorithm for height valve faults, which could detect faults with an accuracy of 97\%.
  Ye et al.~\cite{Ye2022} proposed a detector used the K-means algorithm to design the anchors.
  Tai et al.~\cite{Tai} abandoned the traditional bounding box and made predictions directly on the feature map, using transposed convolution and skipped links to obtain a larger sensory field.
  However, train fault detection is limited by hardware resources in practical scenarios. Meanwhile, the size and the length-width ratio of key components in trains are quite different. Under the premise of low calculation, the above methods are unable to resolve the impact of part size differences on detection.
  
  \subsection{Anchor-free Detectors}
  \begin{figure*}[t]
    \begin{center}
        \centering
        \includegraphics[width=6in]{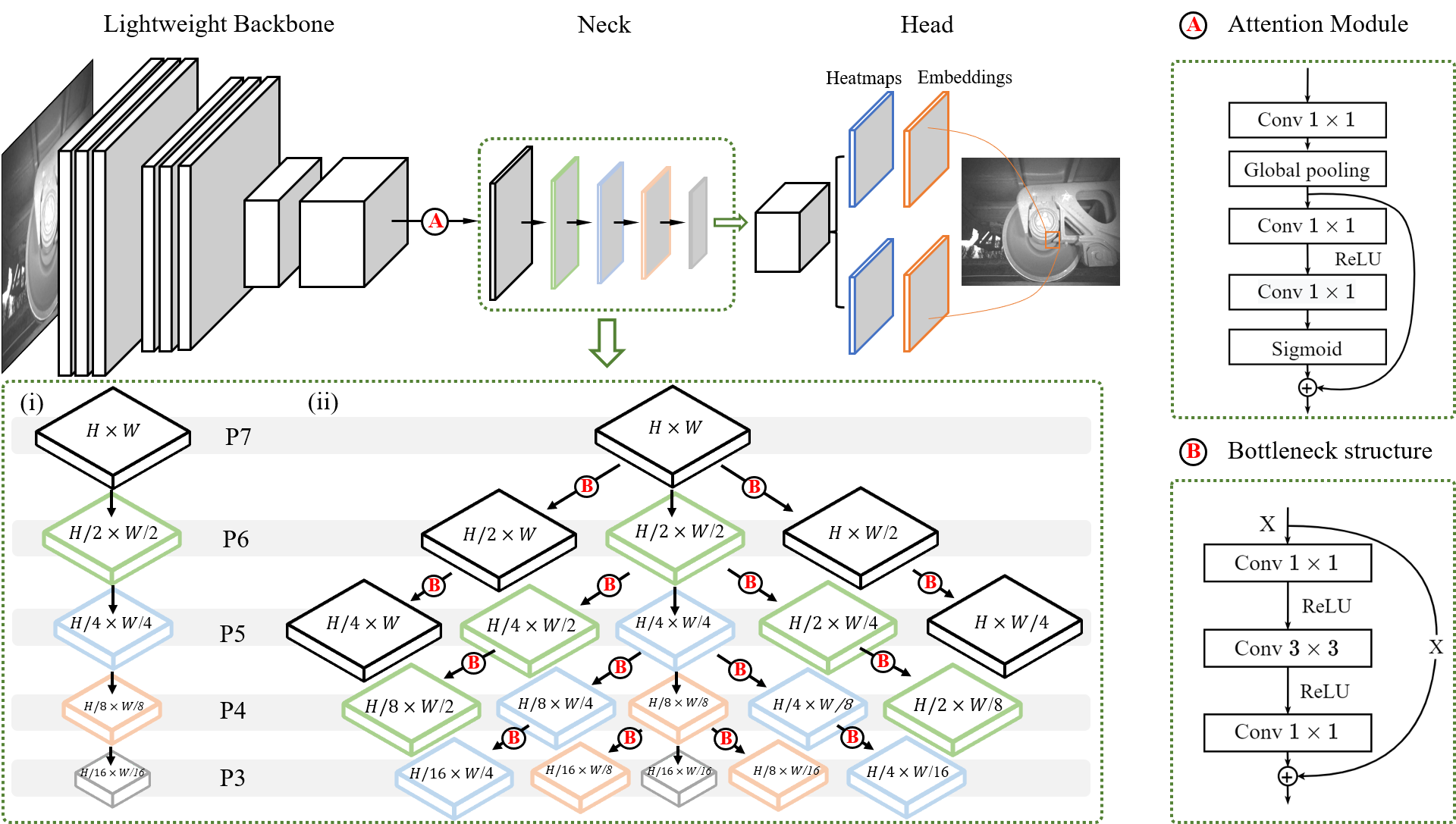}
        \caption{Overall network diagram of our proposed method. This network consists of a lightweight backbone, a neck for multi-scale parts modeling, and a head that uses corners instead of anchors for localization and regression. (i) The traditional feature pyramid network. (ii) The multi-scale feature pyramid (MFP) network used in the neck. The MFP is used to model multi-scale fault features and input the extracted information into the head. We calculate four different loss functions in the head to achieve anchor-free fault detection, thus avoiding anchor-related computations. Attention module is used to model the correlation between channels. Bottleneck structure is used to reduce the amount of parameters of the network.}
        \label{fig2}
    \end{center}
  \end{figure*}

  Anchor-free detectors did not use traditional anchor boxes for detection, thus avoiding related computations to a certain extent. Two-stage detectors, such as RepPoints~\cite{RepPoints}, used points instead of bounding boxes for high-quality localization. Since the one-stage detectors did not require additional RPN for the region proposal, the network structure was simple, and many one-stage anchor-free detectors have appeared recently. For example, CentripetalNet~\cite{CentripetalNet} and CornerNet~\cite{CornerNet} used corners instead of anchors to form bounding boxes. Similar to fully convolutional one-stage (FCOS)~\cite{FCOS} object detection, FoveaBox~\cite{FoveaBox} eliminated the predefined set of anchor boxes and directly predicted the position and boundary of the object. Both AutoAssign~\cite{Autoassign} and adaptive training sample selection (ATSS)~\cite{ATSS} improved the accuracy of anchor-free detectors by automatically distinguishing positive and negative samples. YOLOX~\cite{YOLOX} switched you only look once (YOLO) to anchor free, with decoupled head and advanced label assigning strategy to improve accuracy and speed further.
  Some one-stage detectors removed the non-maximum suppression (NMS) in the post-processing and realized end-to-end detection, such as Libra R-CNN~\cite{Balanced}, CenterNet~\cite{Object}, Deformable DETR~\cite{Deformable_DETR}, Sparse R-CNN~\cite{Sparse_R-CNN}, and OneNet~\cite{OneNet}. These detectors tended to be smaller in model size and required fewer hardware resources during training and testing, but at the expense of accuracy.
   
  \section{DETECTION ALGORITHM}
  \label{Methods}
  We propose a lightweight anchor-free detector using a multi-scale feature pyramid network as a neck to solve the detection problem of different scale parts in trains. As shown in Fig.~\ref{fig2}, this network consists of a lightweight backbone suitable for feature extraction of typical train faults, a feature pyramid network for multi-scale feature modelling, and the head using an anchor-free method for localization and regression. 

  \subsection{Lightweight Backbone}
  Train fault detection is usually carried out in the field environment, with rigorous computing resource requirements. In this case, the lightweight of the network is essential. The backbone is mainly used for feature extraction, and it is the structure with the most hidden layers in the whole network, which mainly determines the model size and amount of network calculation. Therefore, we employ a lightweight backbone to extract train fault features, namely RFDNet~\cite{TII}.
  
  RFDNet uses the fire module, robust to the illumination unevenness of faulty images and has fewer parameters and computation. In the expansion layer, RFDNet deletes the original $1\times1$ convolutional layer, adjusts the overall structure to a streamline. In addition, RFDNet replaces all $3\times3$ convolutions in the expansion layer with depthwise separable convolutions. The number of layers in the network can be made deeper under the same amount of parameters. 
  
  In general, when a convolutional layer has $m$ convolution kernels of size $n \times n$, and each convolution kernel contains $c$ channels, then the parameter $N$ of this convolutional layer can be calculated as $N = m \times n^2 \times c$. For depthwise separable convolution, one operation can be divided into two steps. First, a multi-channel feature map is convolved in a two-dimensional plane. Then the output feature map is convolved through the $1\times1$ convolution to realize the data association between different channels. For the same image, the parameter $N'$ of the depthwise separable convolution can be calculated as $N' = m \times n^2 \times 1 + m \times 1 \times 1 \times c$. As attributes are extracted, depthwise separable convolution can save more parameters.

  \subsection{Multis-cale Feature Pyramid}
  To solve the detection problem caused by scale variation, we use matrix networks in the neck. As depicted in Fig.~\ref{fig2}, the central axis of the matrix can be regarded as a traditional feature pyramid network (FPN). Furthermore, a particular rectangular layer is generated to the left and right in each layer of FPN. The width and height are reduced to half of the diagonal layer in the left and right layers, respectively. With the deepening of the FPN, 19 layers with different scales are developed. Objects of different aspect ratios are assigned to rectangular layers with similar scales to improve accuracy.
    
  Matrix networks are operated by $3 \times3$ convolutions of different strides to generate rectangular layers of different scales. A large amount of computation is performed during these processes, increasing the memory usage and model size. To realize the deployment of the CNN-based model in the wild, we need to lighten this structure. We replace all $3 \times3$ convolutions with the bottleneck module, as shown in Fig.~\ref{fig2}. The first $1 \times1$ convolution reduces the number of channels in the input feature maps from 96 to 48, and the last $1 \times1$ convolution restores channels to 96. In this way, the computation of $3 \times3$ convolutions to change the aspect ratio of the rectangular layer is significantly reduced. Specifically, the number of input channels is $C_{in}$, the number of output channels is $C_{out}$, and the size of the convolution kernel is $K \times K$. The parameter $P$ can be calculated as $P=C_{in} \times K^2 \times C_{out}$. The parameter amount of the original module during the convolution operation is 82944. In contrast, the parameter amount after using the bottleneck structure is 29952 (a reduction of 63.89\%). Moreover, the model size is also reduced, as demonstrated in Section~\ref{Ablation_Study}. After replacing all $3 \times3$ convolutions, this module is named multi-scale feature pyramid network (MFP).

  In addition, the number of channels is adjusted when the MFP receives the feature maps from the backbone. 
  To minimize the loss of information, we add a channel attention module where the MFP is connected to the backbone. As shown in Fig.~\ref{fig2}, this module obtains the global features of each channel through the squeeze operation. And then, it performs the excitation operation on these features. The network can learn the relationship and weight between different channels and apply the weight to the feature map in this process. The model pays more attention to the channel with a large amount of information and ignores some channels with low weight. We will verify the effectiveness of the bottleneck structure and attention module in Section~\ref{Ablation_Study}.

  \subsection{Loss Function}
  We incorporate an anchor-free detector into the framework to avoid computations about anchor generation. In the head, the feature maps processed by the neck are first to develop heatmaps for predicting the upper left and lower right corners of the object. The traditional way of creating heatmaps by CornerNet~\cite{CornerNet} results in much loss of details and inaccurate predictions at the corners of feature maps. However, we provide corresponding prediction outputs for objects of different scales in the neck, ensuring that the aspect ratio of the output feature map is within a certain range. Such an approach can avoid using the corner pooling in the head, which leads to the loss of detailed information.
  
  Furthermore, the pixels occupied in the acquired image are much smaller than the background due to the small scale of the key components of the train. As a result of the imbalance between the foreground and background classes, a large number of negative samples dominate the training process, affecting the accuracy. We utilize focal loss to modulate the imbalanced classes in the predicted class, which is calculated as follows:
  \begin{equation}
      {L_{fl}} = \left\{ \begin{array}{l}
          - \alpha {(1 - y)^\beta }\log y \begin{array}{*{20}{c}}
         {\begin{array}{*{20}{c}}
         {}&{}
         \end{array}}&{x = 1}
         \end{array}\\
          - (1 - \alpha ){y^\beta }\log (1 - y) \begin{array}{*{20}{c}}
         {}&{x = 0}
         \end{array}
         \end{array} \right.,
  \end{equation}
  where $x$ is the true value of the label, $y\in [0,1]$ is the output of the activation function. For positive samples, the larger the predicted probability, the smaller the calculated loss, and vice versa for negative samples. $\alpha$ and $\beta$ are balance factors, $\alpha$ is generally taken as 0.25, which is mainly used to balance the problem of an uneven proportion of positive and negative samples. The value of $\beta$ is usually taken as 2, which is primarily used to adjust the rate of weight reduction.
  
  For the optimization of the offset in the corner regression and the problem of matching the corners in the center point regression, we introduce the smooth L1 loss~\cite{Fast_r-cnn} function to optimize the parameters, and the calculation method is
  \begin{equation}
      {L_{SL{1_i}}} = \left\{ \begin{array}{l}
          0.5{\phi _i}^2\begin{array}{*{20}{c}}
          {\begin{array}{*{20}{c}}
          {}&{}
          \end{array}}&{\left| {{\phi _i}} \right| < 1}
          \end{array}\\
          \left| {{\phi _i}} \right| - 0.5\begin{array}{*{20}{c}}
          {}&{otherwise}
          \end{array}
          \end{array} \right.,
  \end{equation}
  \begin{equation}
      {\phi _i} = \left( {\frac{{{x_i}}}{\delta } - {{\left( {\frac{{{x_i}}}{\delta }} \right)}^\prime },\frac{{{y_i}}}{\delta } - {{\left( {\frac{{{y_i}}}{\delta }} \right)}^\prime }} \right),
  \end{equation}
  where $\phi _i$ represents the offset between the predicted location and the ground-truth location, $\delta$ is the downsampling factor. We can fine-tune the position of the corner by predicting the displacement offset through smooth L1 loss. In the final calculation of the loss, we add all $L_{SL{1_i}}$ to optimize the parameters.
  
  Among several corners predicted for the same object, we apply pull loss to close the corners that belong to the same object and use push loss to move away from the corners that do not belong to the same object~\cite{CornerNet}. When the embeddings corresponding to the upper left and lower right corners of a target object are $e_{ti}$ and $e_{bi}$, the loss function is
  \begin{equation}
      {L_{pull}} = {({e_{ti}} - {e_k})^2} + {({e_{bi}} - {e_k})^2},
  \end{equation}
  \begin{equation}
      {L_{push}} = \max (0,1 - \left| {{e_k} - {e_j}} \right|),
  \end{equation}
  where $e_k$ is the mean of $e_{ti}$ and $e_{bi}$, the total loss function is:
  \begin{equation}
    \label{Eq_loss}
      \begin{array}{l}
          Loss = {L_{fl}} + \frac{1}{n}\sum\limits_{k = 1}^n {{L_{SL{1_i}}}} \\
          \begin{array}{*{20}{c}}
          {}&{}&{ + \lambda (\frac{1}{n}\sum\limits_{k = 1}^n {{L_{pull}}}  + \frac{1}{{n(n - 1)}}\sum\limits_{k = 1}^n {\sum\limits_{j = 1}^n {{L_{push}}} } ),}
          \end{array}
          \end{array}
  \end{equation}
  where $\lambda$ represents the weight of $L_{pull}$ and $L_{push}$, which is generally set to 0.1 in the experiments.

  \section{EXPERIMENTS}
  \label{EXPERIMENTS}
  This section introduces the hardware system for train fault detection, the datasets and the evaluation criteria used for the experiments. And then, we analyze the entire network, detailing it in terms of backbone, neck, loss function, and hyperparameters. Finally, our proposed network is compared with state-of-the-art methods.

  \begin{table*}[t]
    \footnotesize
    \renewcommand\arraystretch{1.5}
    \caption{Comparison of anchor-based and anchor-free methods on the bogie block keys dataset, and comparison of different backbones.}
     \label{Backbone}
    \begin{center}
    \begin{tabular}{lcccccc}
    \toprule
    \textbf{Method} &
      \textbf{Backbone} &
      \textbf{CDR(\%)$\uparrow$} &
      \textbf{FDR(\%)$\downarrow$} &
      \textbf{MDR(\%)$\downarrow$} &
      \textbf{\begin{tabular}[c]{@{}c@{}}Memory\\ Usage(MB)\end{tabular}} &
      \textbf{\begin{tabular}[c]{@{}c@{}}Model\\ Size(MB)\end{tabular}} \\ \midrule
    Anchor-based & ResNet-50 & 97.17 & 2.66 & 0.17 & 2625 & 165.2 \\
    Anchor-free  & ResNet-50 & 96.89 & 2.97 & 0.14 & 1687 & 193.6 \\
    Anchor-based  & RFDNet   & 98.41 & 1.04 & 0.55 & 1721 & 73.1 \\
    Anchor-free  & RFDNet   & 98.96 & 0.90 & 0.14 & 1533 & 101.5 \\ \bottomrule
    \end{tabular}
    \end{center}
    \end{table*}

    \begin{table}[t]
        \footnotesize
        \renewcommand\arraystretch{1.5}
        \caption{Compare the performance of multi-scale feature pyramid with different layers and traditional feature pyramid network on the bogie block keys dataset, using the anchor-free method and RFDNet as the backbone.}
         \label{MFP_table}
        \begin{center}
        \begin{tabular}{lcccccc}
        \toprule
        \textbf{Layer Range} &
        \textbf{\begin{tabular}[c]{@{}c@{}}CDR\\ (\%)$\uparrow$\end{tabular}} &
        \textbf{\begin{tabular}[c]{@{}c@{}}FDR\\ (\%)$\downarrow$\end{tabular}} &
        \textbf{\begin{tabular}[c]{@{}c@{}}MDR\\ (\%)$\uparrow$\end{tabular}} &
        \textbf{\begin{tabular}[c]{@{}c@{}}Memory\\ Usage(MB)\end{tabular}} \\  \midrule
        FPN               & 95.41 & 4.18  & 0.41  & 5409  \\
        (P3-P7) 19 Layers & 98.96 & 0.90  & 0.14  & 8419  \\
        (P4-P7) 14 Layers & 99.31 & 0.66  & 0.03  & 8329  \\
        (P5-P7) 9 Layers  & 98.62 & 1.35  & 0.03  & 7803  \\
        (P6-P7) 4 Layers  & 98.14 & 1.79  & 0.07  & 6911  \\ \bottomrule
        \end{tabular}
        \end{center}
    \end{table}
  
  \subsection{Hardware System}
  The hardware system for train fault detection in the actual scene is shown in Fig.~\ref{exp}. The left side is an acquisition system for train fault images, consisting of an array of high-speed cameras and flashing lights. The system is placed in the middle and on both sides of the track to capture images of critical parts in trains for detection. It can be seen from Fig.~\ref{exp} that the comparison of normal and fault images of key components of the train, in which the size and scale of the faulty components are quite different.
  
  Since those components are located in crucial positions such as train bogies, there are many other parts around, which are easy to cause occlusion during filming. Besides, there is a lot of dust and debris near the wheels, and the lighting in the field is uneven, resulting in blurred images. In addition, the dimensions and shapes of these parts are quite different. These factors lead to an increase in the difficulty of detection.

  \subsection{Experimental Setup}
  \subsubsection{Implementation Details}
  Our method is trained via backpropagation and the Adam optimizer. The hyperparameter settings are the same for a fair comparison in all experiments. We set the initial learning rate to $5e^{-5}$ and cut it by 1/10 after 60K iterations for a total of 70K iterations. We use a batch size of 8 for all ablation studies, and the input images were cropped to a size of $512 \times512$ during training. It is worth noting that we fine-tune the hyperparameters of each model to make the best performance on our hardware devices. All experiments are performed on a PC with an Intel Core i9-9900K 3.60 GHz CPU, 64G RAM and a single NVIDIA GTX 2080Ti GPU.

  \begin{figure}[t]
    \begin{center}
        \centering
        \includegraphics[width=3in]{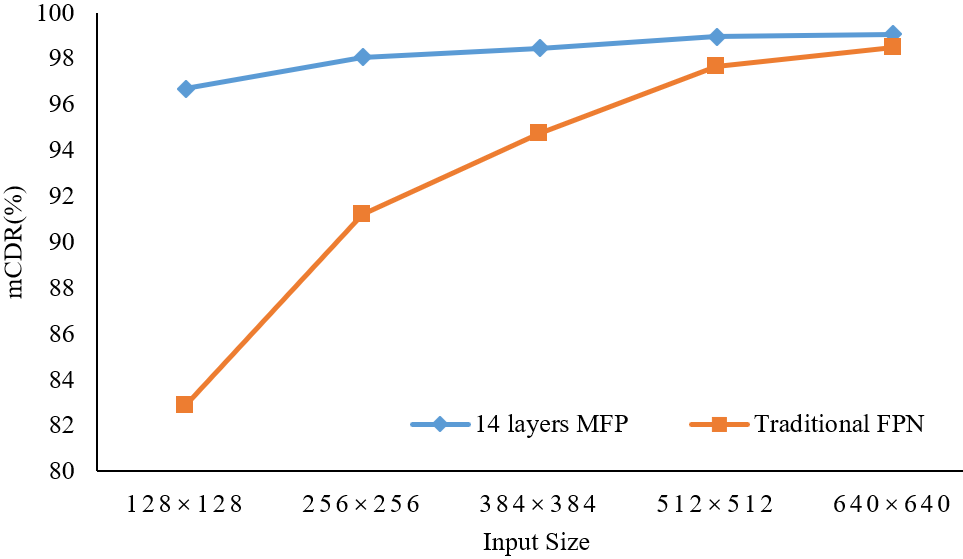}
        \caption{Use input images of different sizes to explore the robustness of the network to scale changes.}
        \label{zhexian}
    \end{center}
  \end{figure}  
    
  \subsubsection{Datasets}
  We focus on fault detection of multi-scale key components of the train braking system, for which we created four datasets of typical component faults using annotation tools. The images for these four datasets come from bogie block keys, dust collectors, cut-out cock and fastening bolts on the brake beam, respectively.
  
  \begin{itemize}
      \item{\textbf{Bogie Block Key}: We built a dataset containing 8337 images to detect bogie block key faults in trains. Among them, 5440 images are used for training, and 2897 images are used for testing. Bogie block key is usually located at the connection between the train wheel axle and bogie, which is used to prevent the separation of the wheel from the bogie. 
      }
      \item{\textbf{Dust Collector}: It is mainly used to filter out the impurities in the air and ensure the purity of the air entering the cylinder. We use 815 images and 850 images for training and testing, respectively. 
      }
      \item{\textbf{Cut-out cock}: It is the critical component used to cut off the air from the central air tank to the brake pipe and is usually located near the dust collector. We use 815 images and 850 images for training and testing, respectively.
      }
      \item{\textbf{Fastening Bolt on Brake Beam}: It is mainly used for the fastening of brake beams in train braking systems. The solid horizontal force can easily cause the fastening bolts to break and fall off during braking. We use 1724 images and 1902 images for training and testing, respectively.
      }
  \end{itemize}

  \begin{table*}[t]
    \footnotesize
    \renewcommand\arraystretch{1.5}
    \caption{Add bottleneck structure and attention module to the network for lightweight and accuracy improvement.}
     \label{lightweight}
    \begin{center}
    \begin{tabular}{cccccccc}
    \toprule
    \multirow{2}{*}{\textbf{\begin{tabular}[c]{@{}c@{}}Bottleneck\\ Structure\end{tabular}}} &
      \multirow{2}{*}{\textbf{\begin{tabular}[c]{@{}c@{}}Attention\\ Module\end{tabular}}} &
      \multirow{2}{*}{\textbf{CDR(\%)$\uparrow$}} &
      \multirow{2}{*}{\textbf{FDR(\%)$\downarrow$}} &
      \multirow{2}{*}{\textbf{MDR(\%)$\downarrow$}} &
      \multicolumn{2}{c}{\textbf{Memory (MB)}} &
      \multirow{2}{*}{\textbf{\begin{tabular}[c]{@{}c@{}}Model\\ Size(MB)\end{tabular}}}  \\ \cline{6-7}
      &   &       &      &      & \textbf{train} & \textbf{test} &       \\ \midrule
      &   & 99.31 & 0.66 & 0.03 & 8329           & 1533          & 101.5  \\
      & \checkmark & 97.51 & 2.04 & 0.45 & 8567           & 1533          & 101.6 \\
      \checkmark &   & 99.00 & 1.00 & 0.00 & 8605           & 1481          & 79.6  \\
      \checkmark & \checkmark & 99.72 & 0.21 & 0.07 & 8645           & 1481          & 79.7\\ \bottomrule
    \end{tabular}
    \end{center}
    \end{table*}

  \subsubsection{Evaluation Metrics}
  To validate the effectiveness of our network, we propose six evaluation metrics, which are false detection rate (FDR), correct detection rate (CDR), missed detection rate (MDR), training memory usage, testing memory usage and model size. The FDR, CDR, and MDR are indicators used to evaluate the accuracy of the detector. They are defined as follows: there are $a$ images identified as fault images, and $c$ images are identified as normal images in the test set. Furthermore, $b$ images in $a$ and $d$ images in $c$ are detected incorrectly, respectively. $m$ represents the number of faulty images in the dataset, $n$ represents the number of remaining images in the dataset, then:
  \begin{equation}
      \label{deqn_ex1}
      FDR = \frac{b}{m + n},\  CDR = \frac{a+c}{m+n},\  MDR = \frac{d}{m+n}.
  \end{equation}
  Model size intuitively reflects the computational cost of the model, and the memory usage of training and testing reflects the dependence of the model on the hardware system. We expect these two metrics to be as low as possible for the same detection accuracy. We calculate the average of FDR, CDR and MDR, i.e. mFDR, mCDR and mMDR to evaluate the effectiveness of the model when using four datasets.

  \begin{table}[t]
    \footnotesize
      \renewcommand\arraystretch{1.5}
      \caption{The performance of MFP structure with different number of channels on the bogie block keys dataset. NC means cannot converge.}
       \label{channels}
      \begin{center}
    \begin{tabular}{ccccccccc}
      \toprule
  \multirow{2}{*}{\textbf{\begin{tabular}[c]{@{}c@{}}Number of\\ Channels\end{tabular}}} &
    \multirow{2}{*}{\textbf{\begin{tabular}[c]{@{}c@{}}CDR\\ (\%)$\uparrow$\end{tabular}}} &
    \multirow{2}{*}{\textbf{\begin{tabular}[c]{@{}c@{}}FDR\\ (\%)$\downarrow$\end{tabular}}} &
    \multirow{2}{*}{\textbf{\begin{tabular}[c]{@{}c@{}}MDR\\ (\%)$\downarrow$\end{tabular}}} &
    \multicolumn{2}{c}{\textbf{Memory (MB)}} &
    \multirow{2}{*}{\textbf{\begin{tabular}[c]{@{}c@{}}Model\\ Size (MB)\end{tabular}}} \\ \cline{5-6}
        &       &      &      & \textbf{train} & \textbf{test} &      \\ \midrule
  256   & 99.72 & 0.21 & 0.07 & 8645           & 1481          & 79.7 \\ 
  192   & 97.83 & 1.69 & 0.48 & 5461           & 1677          & 52.6 \\
  128   & 98.72 & 0.79 & 0.48 & 4645           & 1671          & 30.5 \\
  96    & 99.21 & 0.66 & 0.14 & 4293           & 1653          & 22.5 \\
  64    & NC   & NC  & NC  & 3819           & 1641          & 16.5 \\ \bottomrule
  \end{tabular}
  \end{center}
  \end{table}

  \subsection{Ablation Studies}
  \label{Ablation_Study}
  \subsubsection{Overall Network}

  We explore the overall structure of the network to select the most suitable method for fault detection of multi-scale key components in freight trains. We use MatrixNets as the baseline to evaluate the performance of anchor-based and anchor-free methods on the bogie block key dataset.
  As shown in Table~\ref{Backbone}, the anchor-free method achieves minor memory usage, which is extremely important for fault detection with limited computing resources.
  As the feature extraction part of the entire network, the backbone undertakes most of the network computations and largely determines the model size, memory usage, and detection accuracy. 
  To optimize the above metrics for anchor-free detectors under low memory usage, we replace the backbone from ResNet-50 with RFDNet~\cite{TII}, a more lightweight backbone. As can be seen from the comparison in Table~\ref{Backbone}, the model size after using RFDNet is reduced from 193.6 MB to 101.5 MB, which is better than the two methods used in the baseline. 
  Under the premise of small model size and low memory usage, the method using RFDNet as the backbone is even better than the formers in terms of accuracy. These results show that the feature extraction effect of RFDNet for typical train components is better than ResNet-50, and it can provide more helpful feature maps for subsequent networks.
  
  \subsubsection{Multi-scale Feature Pyramid Design}
  \label{MFP}  
      To verify the effectiveness of the MFP, we compare the performance of the traditional FPN with the MFP on the bogie block keys dataset. Meantime, the aspect ratios of the critical parts have a specific range. As a result, we use the MFP of 19 layers as the baseline and gradually reduce the number of MFP layers to find the best structure most suitable for the feature extraction of train faults. As shown in Table~\ref{MFP_table}, we use the RFDNet as the backbone and the anchor-free method to conduct experiments. 
      
      The CDR of FPN on the bogie block keys dataset is only 95.41\%, which is worse than MFP of any structure. This result verifies the effectiveness of MFP, indicating that a reasonable MFP structure can predict faults at corresponding scales by outputting feature maps of different scales. Among MFPs with different layers, the 14 layers MFP has the highest accuracy, reaching 99.31\%. 
      Oversampling under 19 layers MFP loses details significantly. On the other hand, 9 layers and 4 layers MFP outputs fewer feature maps to match components with different aspect ratios. Therefore, a reasonable selection of the layer range can effectively avoid the accuracy decreases caused by oversampling or undersampling.

      It should be highlighted that we explore the best MFP structure by suppressing the outputs of some levels. The relevant properties of the suppressed layers are not trained, but they are still retained. As a result, the memory usage (1533 MB) and model size (101.5 MB) of various layers are consistent during testing.
        \begin{table}[t]
          \footnotesize
              \renewcommand\arraystretch{1.5}
              \caption{Validation of $L_{pull}$ and $L_{push}$ in the training process. NC means cannot converge.}
               \label{loss_table}
              \begin{center}
          \begin{tabular}{ccccccc}
            \toprule
          \textbf{$\lambda$}     & 0    & 0.1   & 0.3   & 0.5   & 0.8   & 1     \\ \midrule
          \textbf{CDR(\%)$\uparrow$} & NC & 99.21 & 98.72 & 97.52 & 97.93 & 98.10 \\
          \textbf{FDR(\%)$\downarrow$} & NC & 0.66  & 0.93  & 2.00  & 1.52  & 1.80  \\
          \textbf{MDR(\%)$\downarrow$}  & NC & 0.14  & 0.35  & 0.48  & 0.55  & 0.10 \\ \bottomrule
          \end{tabular}
        \end{center}
          \end{table}

          \begin{table*}[t]
            \footnotesize
            \renewcommand\arraystretch{1.5}
            \caption{Comparison of the detection results on four typical faults with state-of-the-art methods.}
            \label{SOTA}
            \begin{center}
            \begin{tabular}{lcccccccc}
            \toprule
            \multirow{2}{*}{\textbf{Methods}} &
            \multirow{2}{*}{\textbf{Backbone}} &
            \multirow{2}{*}{\textbf{\begin{tabular}[c]{@{}c@{}}mCDR$\uparrow$\\ (\%)\end{tabular}}} &
            \multirow{2}{*}{\textbf{\begin{tabular}[c]{@{}c@{}}mMDR$\downarrow$\\ (\%)\end{tabular}}} &
            \multirow{2}{*}{\textbf{\begin{tabular}[c]{@{}c@{}}mFDR$\downarrow$\\ (\%)\end{tabular}}} &
            \multirow{2}{*}{\textbf{\begin{tabular}[c]{@{}c@{}}Batch\\ size\end{tabular}}} &
              \multicolumn{2}{c}{\textbf{Memory(MB)}} &
              \multirow{2}{*}{\textbf{\begin{tabular}[c]{@{}c@{}}Model\\      size(MB)\end{tabular}}} \\ \cline{7-8}
              &                  &       &      &   &    & \textbf{Train*} & \textbf{Test}     \\ \midrule
              HOG+Adaboost+SVM$\dagger$~\cite{TIM}        & --        &94.39 & 1.35 & 4.26 & -- & --  & --       & 0.12      \\
              FAMRF+EHF$\dagger$~\cite{OE}          	   & --       &94.96 &1.00  &4.04 & -- & --  & --       & 0.11      \\
              Cascade detector (LBP)$\dagger$~\cite{OE}   & --       &88.69 &3.77 &7.54 & -- & --  & --       & --     \\
              FTI-FDet**~\cite{TIM}           & VGG16    & 99.27   & 0.52  & 0.21  & 128 & --  & 1823          & 557.3      \\
              Light FTI-FDet**~\cite{TIM}          & VGG16    & 98.91   & 0.49  & 0.61   & 128 & --  & 1533          & 89.7      \\\midrule
              AutoAssign~\cite{Autoassign}      & ResNet-50        & 96.95   & 1.97 & 1.08  & 2 & 1528     & 1275          & 289.2    \\
              CentripetalNet~\cite{CentripetalNet}  & HourglassNet-104 & 95.34   & 3.03 & 1.63 & 2 & 3878    & 3951          & 2469.1   \\
              Cornernet~\cite{CornerNet}       & HourglassNet-104 & 96.42   & 1.26 & 2.32 & 2  & 3643       & 3543          & 2412.4  \\
              FCOS~\cite{FCOS}            & ResNet-50        & 97.53   & 1.99 & 0.48  & 6  & 1311    & \textbf{953}   & 256.1  \\
              FoveaBox~\cite{FoveaBox}        & ResNet-50        & 96.82   & 2.24 & 0.94   & 4 & 1186    & 987      & 289.3    \\
              YOLOX~\cite{YOLOX}           & Modified CSP v5  & 96.64   & 0.49 & 2.87  & 2 & 1932           & 1047          & 650.7    \\
              Libra R-CNN~\cite{Balanced}     & ResNet-50        & 96.53   & 2.66 & 0.81 & 6 & 1614   & 1573       & 332.4    \\
              CenterNet~\cite{Object}       & ResNet-18        & 94.29   & 2.14 & 3.57 & 16  & \textbf{162}    & 1203      & 113.8   \\
              Deformable DETR~\cite{Deformable_DETR} & ResNet-50        & 84.28   & 2.51 & 13.21 & 2 & 2009      & 1131     & 482.1   \\
              Sparse R-CNN~\cite{Sparse_R-CNN}    & ResNet-50        & 90.08   & \textbf{0.38} & 9.54  & 6  & 1538    & 1951   & 1272.6 \\
              OneNet~\cite{OneNet}          & ResNet-50        & 92.82   & 1.51 & 5.67   & 4 & 2200    & 1147     & 355     \\
              MatrixNets~\cite{xNets}          & ResNet-50        & 93.57   & 4.76 & 1.67 & 8 & 1052     & 1687          & 193.6      \\ \midrule
              \multirow{2}{*}{MS FTI-FDet (Ours)}    & ResNet-50    & \textcolor{black}{97.89} & \textcolor{black}{1.06}  & \textcolor{black}{1.05} & \textcolor{black}{8} & \textcolor{black}{891}  & \textcolor{black}{1909}  & \textcolor{black}{114.7}   \\
                      & RFDNet           & \textcolor{black}{\textbf{98.44}} & \textcolor{black}{1.10} & \textcolor{black}{\textbf{0.46}} & \textcolor{black}{8} & \textcolor{black}{537}  & \textcolor{black}{1641} & \textcolor{black}{\textbf{22.5}}  \\ \bottomrule
            \end{tabular}
            \begin{tablenotes}
                * The relevant parameters in each detector are fine-tuned for optimal performance, but the batch size directly affects the memory usage during training. For a fair comparison, the memory usage during training in the table represents the memory consumed by each batch.
        
                ** The methods were experimented with using the publicly available Caffe~\cite{caffe} on a single GTX1080Ti.
        
                $\dagger$ The methods are traditional detectors.
            \end{tablenotes}
            \end{center}
          \end{table*}
  
  \subsubsection{Different Scales}
      To verify the robustness of the network for fault detection of multi-scale parts, we resize the input images. As illustrated in Fig.~\ref{zhexian}, the network uses the anchor-free method and RFDNet as the backbone for experiments. The detection accuracy varies by around 2\% across the five image sizes when using 14 layers (P4-P7) MFP as the neck. However, the detection accuracy is very sensitive to the change of the input image size when using traditional FPN as the neck. These findings suggest that the structural design of the MFP makes the network robust to fault detection of different scales and can solve the influence of the scale variation of the parts on the detection.
  
  \subsubsection{Different Modules}
  \label{Network_Lightweight}
      We add two modules to the network to verify that the bottleneck structure and attention module can reduce the model parameters while improving accuracy. We use RFDNet as the backbone and anchor-free method with 14 layers (P4-P7) MFP for experiments on the bogie block keys dataset. In Table~\ref{lightweight}, the separate addition of the attention module and the bottleneck structure leads to a drop in accuracy. However, the bottleneck structure significantly reduces the model size. The network accuracy and model size are optimized with two structures added simultaneously, making the model more suitable for deployment in scenarios with limited computing resources.

  \subsubsection{\textcolor{black}{Hyperparameter}}
  \label{Hyperparameter}
      To explore the impact of hyper-parameter settings on network performance, we perform ablation experiments on the number of channels in the neck and the weights of the loss function.
  
      Feature information is stored in the channels of the feature map, and it is necessary to reasonably design the number of channels according to the feature complexity of the detected object.
      As shown in Table~\ref{channels}, we sequentially reduce the number of channels in the MFP from 256 to 64, and the model size and detection accuracy fluctuate to varying degrees. When the number of channels is 96, the detection accuracy is only 0.51\% lower than 256, while the model size is 3.5$\times$ smaller. The small model size is extremely important in scenarios of train fault detection where computing resources are limited. As a result, choosing a model with 96 channels is more reasonable for train fault detection. It is worth mentioning that when the number of channels is reduced to 64, the network cannot converge, which shows that only using MFP of 64 channels will significantly lose feature information.
    
      Parameter $\lambda$ in Eq.~\ref{Eq_loss} reflects the importance of $L_{pull}$ and $L_{push}$ in training, and we perform ablation experiments in Table~\ref{loss_table}. When $\lambda = 0$, the network cannot converge, which shows the effectiveness of $L_{pull}$ and $L_{push}$~\cite{CornerNet}. Nevertheless, as $\lambda$ increases, the detection accuracy decreases. As described above, $L_{pull}$ and $L_{push}$ distinguish whether the predicted corners belong to the same object. 
      The weights of $L_{pull}$ and $L_{push}$ are minor because each image in the train datasets only has one target to detect.
  
  \subsection{Comparison with State-of-the-art Methods}
  
      To further highlight the advantages of our proposed multi-scale fault detection network for freight train images (MS FTI-FDet), we compare it with state-of-the-art methods, including fault detectors and anchor-free detectors.
      We conduct experiments on four fault datasets of freight trains to compare accuracy, memory usage, and model size. Moreover, the critical settings of each detector are fine-tuned for optimal performance, and the memory usage during training represents the memory consumed by each batch.
  
      As shown in Table~\ref{SOTA}, the mCDR of MS FTI-FDet is improved by 4.87\% compared to our baseline network MatrixNets. In the meantime, the memory usage during testing is reduced, which is more conducive to model deployment. Furthermore, the model size of MS FTI-FDet is only 22.5 MB, which can reduce the weight of the model while improving the detection accuracy.
      The detection accuracy is improved when using RFDNet as the backbone compared to using ResNet-50, while the model size is reduced by $5\times$. This result demonstrates that RFDNet can extract the features of train faults more efficiently. Compared with other state-of-the-art anchor-free methods, MS FTI-FDet has the highest detection accuracy (98.44\%) and the smallest model size (only 22.5 MB). Our proposed method achieves high-accuracy fault detection with low resource consumption. 
      Compared with the state-of-the-art methods for train fault detection, MS FTI-FDet still has the smallest model size and competitive detection accuracy.

\begin{figure}[t]
    \begin{center}
        \centering
        \includegraphics[width=3.4in]{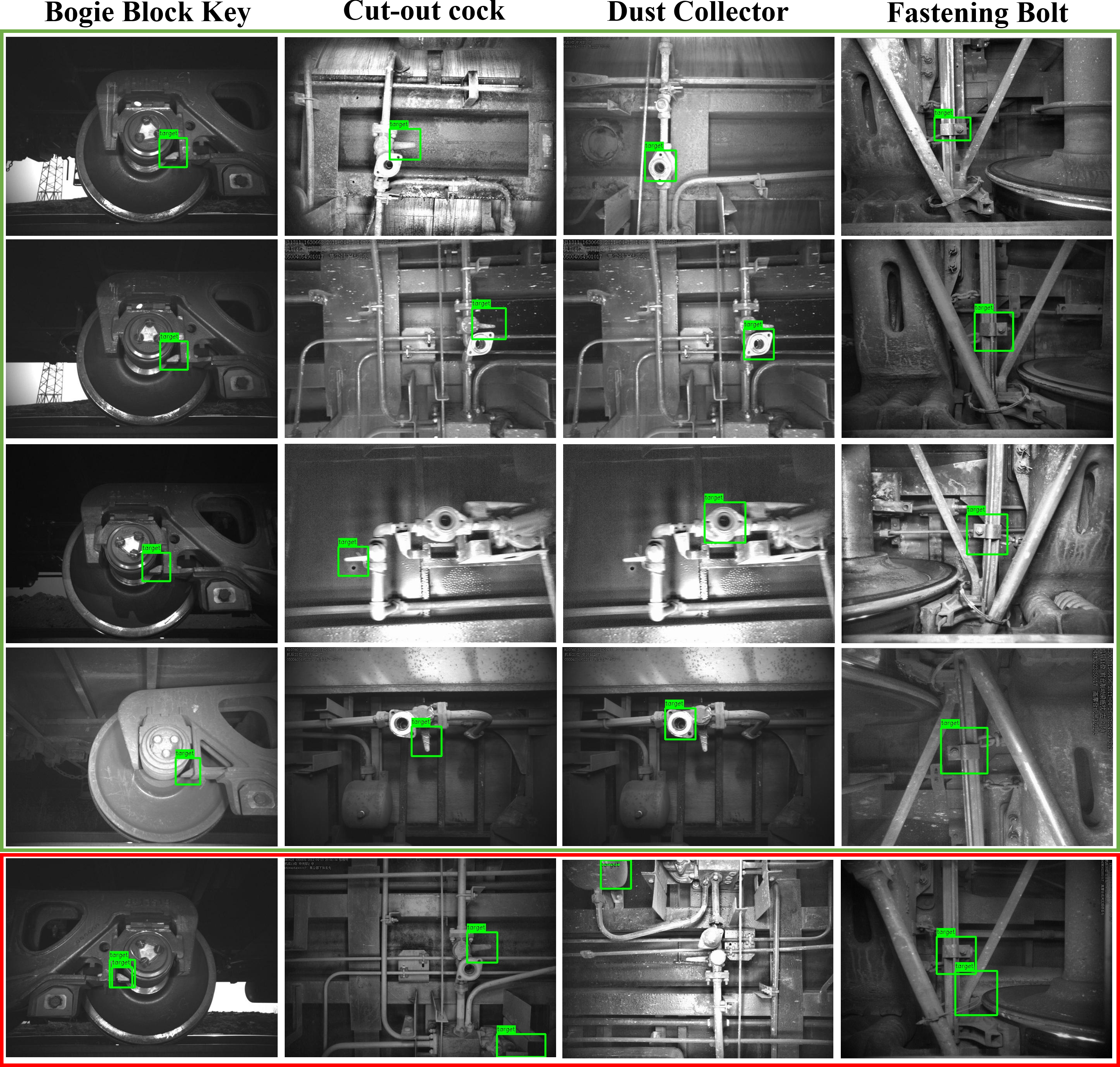}
        \caption{Qualitative results of our method. The first four rows are the correct detection results, and the bottom row is the false results. The robustness of our method to illumination is not satisfactory. We will further expand our datasets to address this problem in the feature.}
        \label{result_png}
    \end{center}
  \end{figure}

      The experimental results in Table~\ref{SOTA} confirm that the scale problem in train fault detection affects the accuracy to a certain extent. 
      MS FTI-FDet surpass all state-of-the-art methods when it comes to solving the scale problem.
      Through our optimization of the network structure, MS FTI-FDet can reduce the demand for hardware resources while ensuring detection accuracy, suitable for vehicle fault detection deployment in field scenarios.
	  We show the detection results in the four datasets in Fig.~\ref{result_png}. It is clear to see that our method has missing and false detections when the images are contaminated or have uneven illumination. The lack of such images in datasets is the critical factor, and we will further expand our datasets to address this problem and improve the generalization ability of the network.
  

  \section{CONCLUSION}
  \label{Results}

  In this article, we propose a lightweight anchor-free framework MS FTI-FDet for the fault detection of different scale parts in trains. The proposed framework consists of a lightweight backbone for feature extraction of train faults, a feature pyramid network for generating rectangular layers of different sizes to map multi-scale components, and a detection head using the anchor-free method for localization and regression. Experiments on four fault datasets indicated that our MS FTI-FDet is robust to fault detection of crucial components at different scales of trains. 
  By solving the scale problem, MS FTI-FDet achieves an accuracy of 98.44\% while significantly reducing the model size, only 22.5 MB, the smallest but most accurate among all state-of-the-art anchor-free detectors. In the future, we will continue to improve our framework to achieve real-time detection and expand the dataset to make the model robust to illumination changes.

  \bibliographystyle{ieeetr}
  \bibliography{refs}

\end{document}